\documentclass[conference]{IEEEtran}
\IEEEoverridecommandlockouts
\usepackage{cite}
\usepackage{amsmath,amssymb,amsfonts}
\usepackage{algorithmic}
\usepackage{graphicx}
\usepackage{textcomp}
\usepackage{xcolor}
\usepackage{tikz}
\usepackage{amsmath}
\usepackage{booktabs}
\usepackage{multirow}
\usepackage{xcolor}
\def\BibTeX{{\rm B\kern-.05em{\sc i\kern-.025em b}\kern-.08em
    T\kern-.1667em\lower.7ex\hbox{E}\kern-.125emX}}

\begin{document}

\title{IMMVP: An Efficient Daytime and Nighttime On-Road Object Detector
}

\author{\IEEEauthorblockN{Cheng-En Wu}
\IEEEauthorblockA{\textit{Institute of Information Science}, \\
\textit{Academia Sinica}, \\
Taipei, Taiwan \\
chengen@iis.sinica.edu.tw}
\and
\IEEEauthorblockN{Yi-Ming Chan}
\IEEEauthorblockA{\textit{Institute of Information Science}, \\
\textit{Academia Sinica}, \\
Taipei, Taiwan \\
yiming@iis.sinica.edu.tw}
\and
\IEEEauthorblockN{Chien-Hung Chen}
\IEEEauthorblockA{\textit{Institute of Information Science}, \\
\textit{Academia Sinica}, \\
Taipei, Taiwan \\
redsword26@gmail.com}
\and
\IEEEauthorblockN{Wen-Cheng Chen}
\IEEEauthorblockA{\textit{Dept. of Computer Science and Information Engineering} \\
\textit{National Cheng Kung University}\\
Tainan, Taiwan \\
dreamfantasy0@gmail.com}
\and
\IEEEauthorblockN{Chu-Song Chen}
\IEEEauthorblockA{\textit{Academia Sinica \& MOST Joint Research Center} \\
\textit{for AI Technology and All Vista Healthcare}\\
Taipei, Taiwan \\
song@iis.sinica.edu.tw}
}

\maketitle

\begin{abstract}
% To detect object in Nvidia Jetson TX2 platform, we take the lightweight model ResNet-18 as the backbone. Then the Feature Pyramid Network generate good feature for each input. With Cascade R-CNN technique, the bounding boxes are iteratively refined for better results. To improve the quality of the classifier, three techniques are used. We define subclasses for day and night time samples. Then we drop similar samples in the training set to prevent overfitting. The high quality outside training samples also help us improve the final results.
It is hard to detect on-road objects under various lighting conditions. To improve the quality of the classifier, three techniques are used. We define subclasses to separate daytime and nighttime samples. Then we skip similar samples in the training set to prevent overfitting. With the help of the outside training samples, the detection accuracy is also improved. To detect objects in an edge device, Nvidia Jetson TX2 platform, we exert the lightweight model ResNet-18 FPN as the backbone feature extractor. The FPN (Feature Pyramid Network) generates good features for detecting objects over various scales. With Cascade R-CNN technique, the bounding boxes are iteratively refined for better results. 
\end{abstract}


\section{Introduction}
Deep leaning has demonstrates its great success on image classification and object recognition.
Recently, various deep convolution neural networks (CNNs) have been proposed for object detection and achieve impressive performance, such as faster R-CNN \cite{ren2016faster}, YOLOv2/v3 \cite{redmon2017yolo9000, Redmon2018YOLOv3AI}, and SSD \cite{liu2016ssd}.
To further improve the detection accuracy, some promising techniques have been studied.
For example, Feature Pyramid Networks \cite{lin2017feature} exploit the intermediate-level features for detecting small objects without a heavy computational burden. 
Cascade R-CNN\cite{cai2018cascade} iteratively applies the process of bounding box regression and classification to sequentially refine the object detection results.
On the other hand, to reduce the computational resource consumed and improve the inference speed, lightweight models such as MobileNet-v2 \cite{sandler2018mobilenetv2} and Pelee \cite{wang2018pelee} have been introduced as well.

The purpose of this work is to design a lightweight model that can perform object detection on the road by using an edge-computing device (e.g.~NVIDIA Jetson TX2).
We target three classes of road object, \textbf{pedestrian}, \textbf{vehicle}, and \textbf{rider}.
To fulfill the goal of the MMSP2019 Embedded Deep Learning Object Detection Model Competition (briefed as MMSP Competition below), we have to design an accurate-enough model (with the mAP at least 0.5), whereas the inference speed of the model is expected to be as faster as possible.

The rest of this paper is organized as follows.
In Section~\ref{design}, we depict the rational of our design and the deep-learning model conducted for efficient object detection on road.
In Section~\ref{experiments}, we present the experimental results on various settings of the object-detection models studied.
Finally, conclusions are given in Section~\ref{conclusion}.

\begin{figure}[t]
  \begin{center}
    \includegraphics[width=0.8\columnwidth]{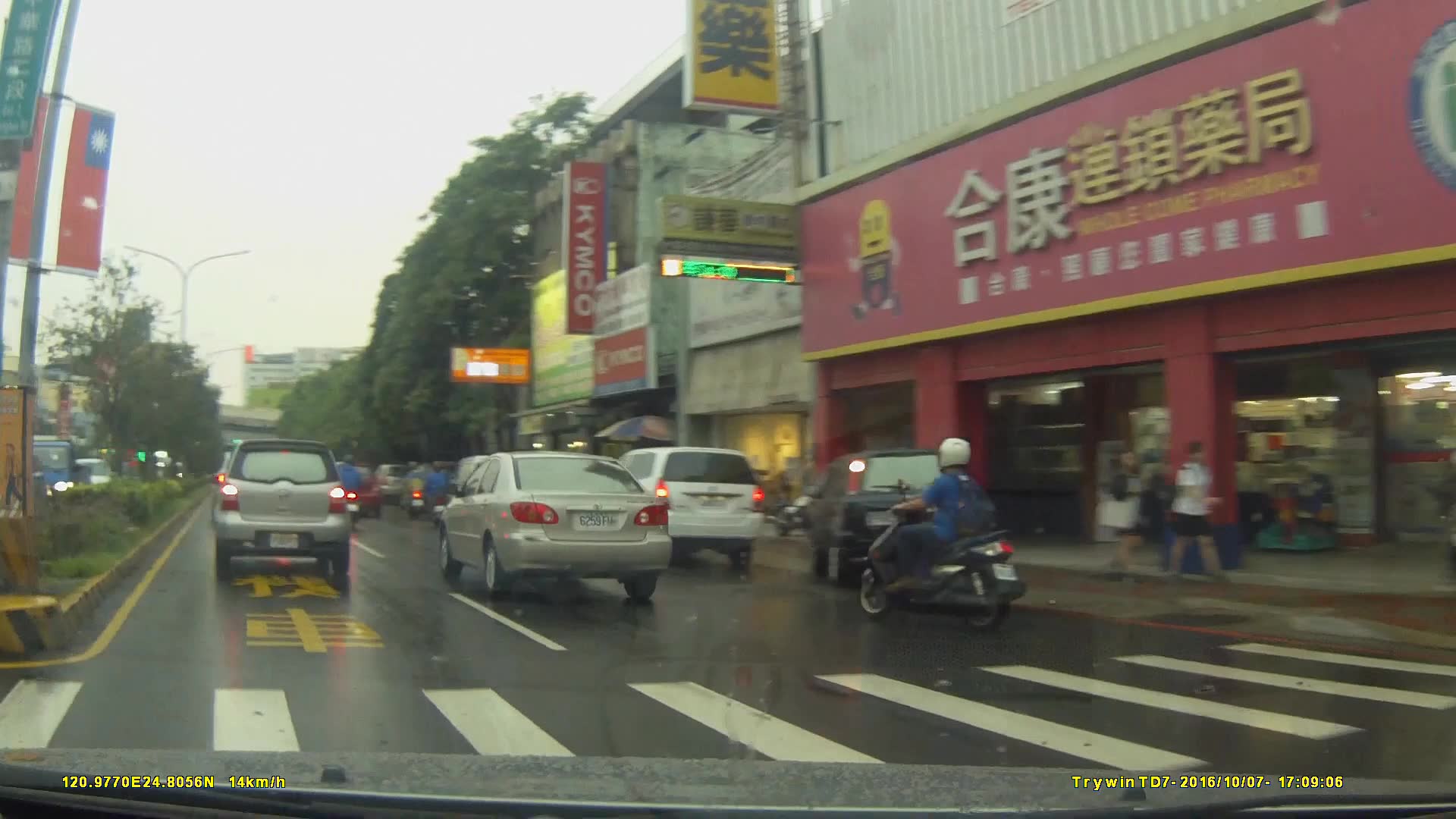}
  \end{center}
  \caption{An example of on road objects in the daytime.}
  \label{fig:day_example}
  \begin{center}
    \includegraphics[width=0.8\columnwidth]{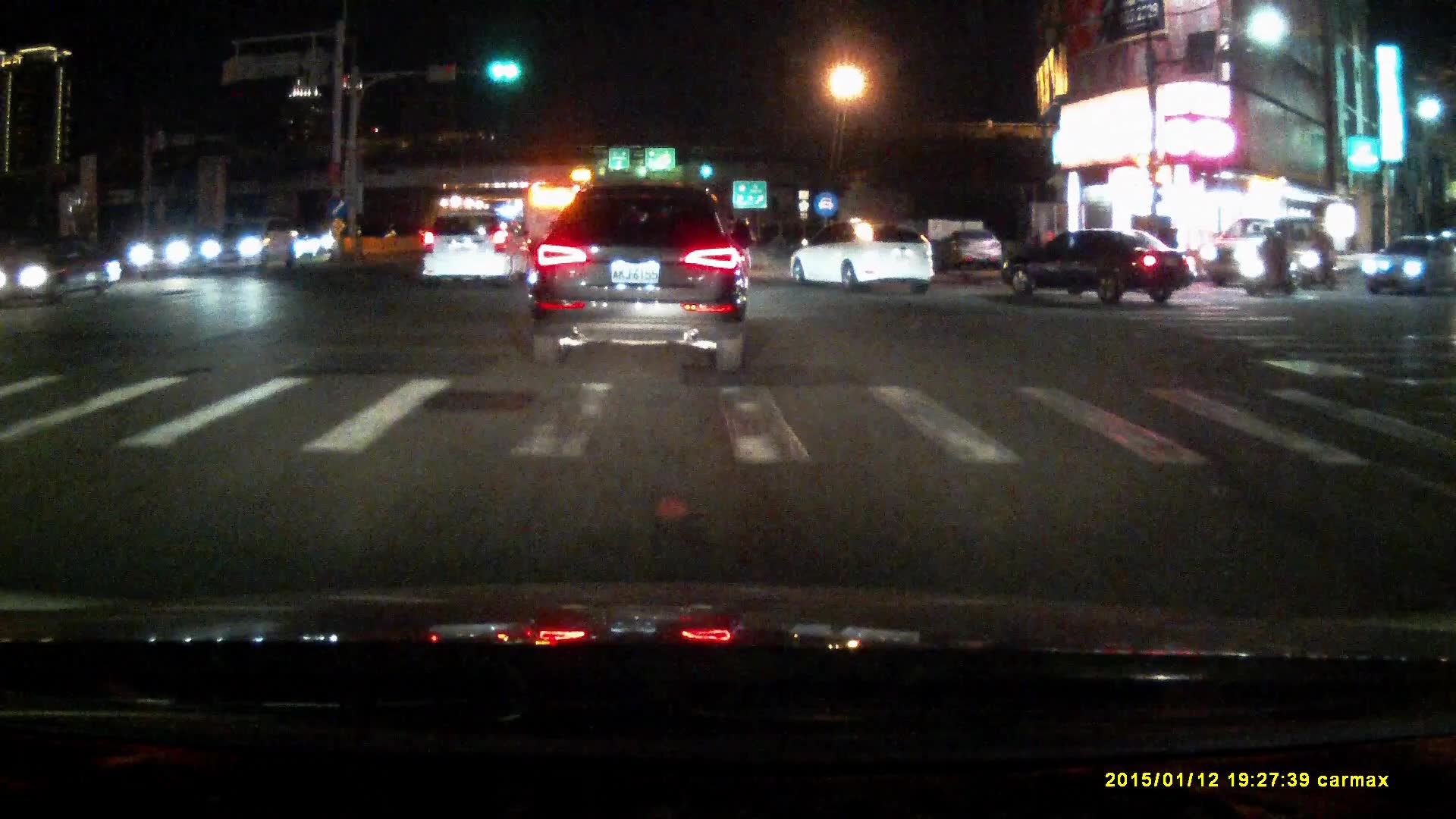}
  \end{center}
  \caption{An example of on road objects in the nighttime.}
  \label{fig:night_result}
\end{figure}

% % Figure
% \begin{figure}[t]
%   \begin{center}
%     \includegraphics[width=0.8\columnwidth]{figures/intro_night.jpg}
%   \end{center}
%   \caption{An example of on road objects in the nighttime.}
%   \label{fig:night_result}
% \end{figure}

% Figure
\begin{figure*}[t]
  \begin{center}
    \includegraphics[width=1.0\textwidth]{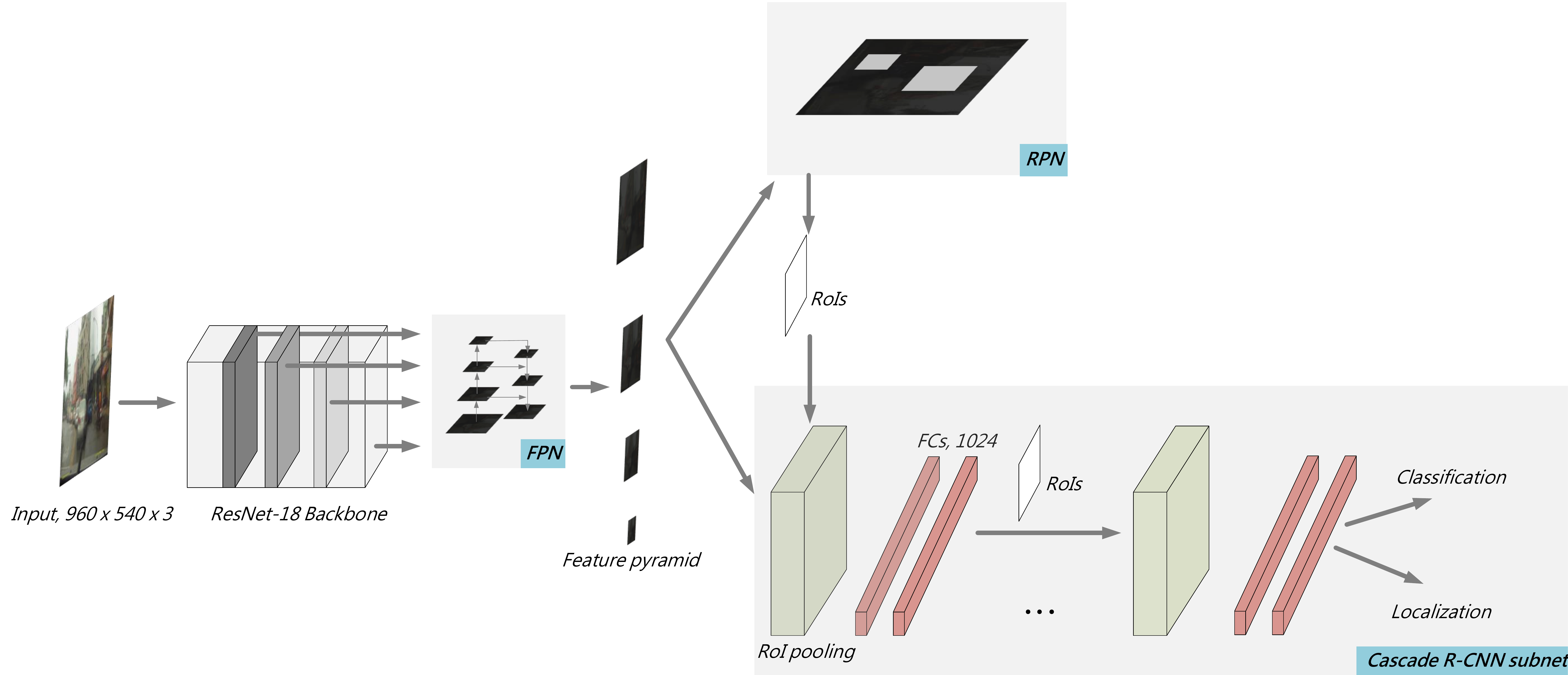}
  \end{center}
  \caption{The overall architecture of the proposed model. To meet the required accuracy of 0.5 mAP with inference efficiency, our model employs 960$\times$540 input resolution. The model is based on Cascade R-CNN \cite{cai2018cascade} but we replace the backbone with ResNet-18 \cite{he2016deep} with Feature Pyramid Network (FPN) \cite{lin2017feature} added.}
  \label{fig:model_arc}
\end{figure*}

%\section{Related Work}
\section{Design Rational and Proposed Object Detector}
\label{design}

In this section, we elaborate on the object-detection model designed for this work.
Our design compromises the inference speed and accuracy.
First, we focus on the development of object-detection models that can achieve a high testing accuracy, as introduced in Section \ref{accuracy}.
Then, to improve the inference efficiency, we reduce the model architecture selected above while maintaining the minimal level of accuracy required, as introduced in Section \ref{efficiency}.

%In this section, we elaborate on our proposed object detection model. Our design mainly focuses on efficiency achieving the required accuracy. In the following two sections, we present methods to increase accuracy in section A and solutions to create a lightweight and efficient model in section B, respectively.

\subsection{Design of High-Accuracy Model}
Our design strategy is to select the model of the highest accuracy from the existing state-of-the-art ones at first and then improve the efficiency of the model. 
Among the existing models, Cascade R-CNN \cite{cai2018cascade} with ResNeXt-101\cite{xie2017aggregated} backbone has the best accuracy on MS COCO dataset \cite{lin2014microsoft}. To further boost the performance, we add Feature Pyramid Network (FPN) \cite{lin2017feature} to the backbone of the Cascade R-CNN model so that features at different scales can be extracted better.
However, this model has achieved merely the mAP of 0.40 on the 1st-stage public testing dataset announced by MMSP Competition, which is much lower than 0.5 (mandatory criterion of mAP). 
Hence, to enhance the accuracy, we adopt three strategies for the improvement in the training stage: sample frames selection, label expansion, and training data increments, as depicted below.

\label{accuracy}
\subsubsection{Sample Frames Selection}
We found that there is much redundancy among the training data provided.
Removing the redundancy can not only save the training time per trial but also increases the accuracy.
Although multiple image sequences are contained in the training data, continuous frames in a sequence have monotonous and repeated information. 
When the data are directly applied to a typical batch-based learning procedure, the resulted models cannot generate comprehensive feature representations. 
To address this issue, we perform uniform sampling for each image sequence. 
We observed that after such pre-processing, a batch of training data owns more representative samples compared to that of utilizing full-sequence images.
According to our experiments, the best accuracy of the model is achieved by uniformly sampling one frame from ten in a sequence.
The training speed is considerably accelerated too.

\subsubsection{Labels Extension}
\begin{figure}[t]
  \begin{center}
    \includegraphics[width=0.8\columnwidth]{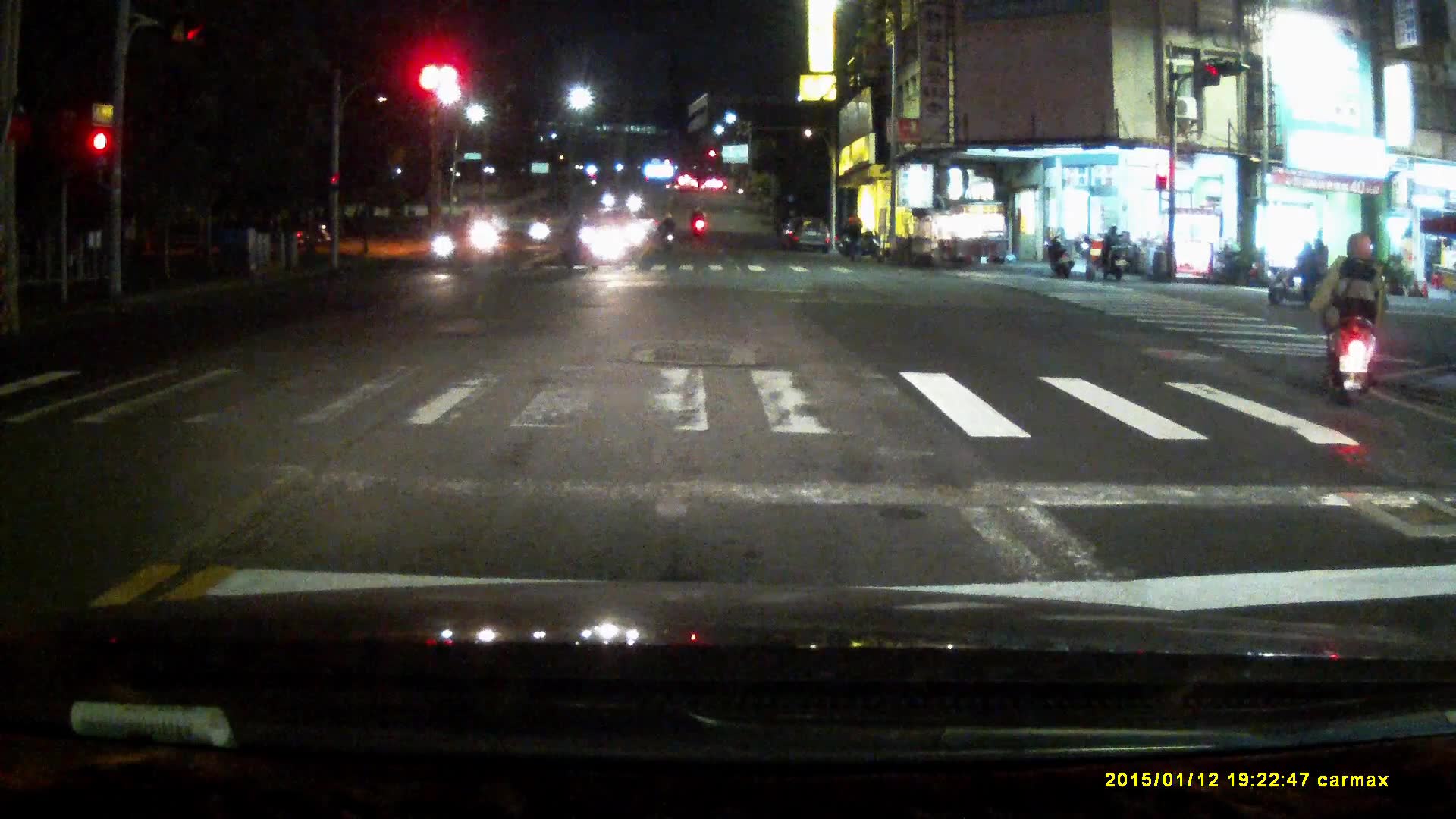}
  \end{center}
  \caption{An example of the vehicles with glare headlights during the nighttime.}
  \label{fig:night_example}
\end{figure}

The task is to detect three types of objects including pedestrian, vehicle, and rider.
The training dataset covers different times of days, including daytime and nighttime.
Since the same object has quite different appearances on daytime and nighttime, it is difficult for a model to learn good feature representations of the objects across the times. 
For example, as shown in Figure~\ref{fig:night_example}, the vehicles during the nighttime are only observable via their headlights, causing the appearance to be highly dissimilar to those that are seen during the daytime.
To address this issue, we propose a label-extension strategy, where six labels (daytime$\_$vehicle, daytime$\_$pedestrian, daytime$\_$rider, night$\_$vehicle, night$\_$pedestrian and night$\_$rider) are used instead of the original three labels (vehicle, pedestrian, rider) in the training stage.
When calculating the accuracy in the training and inference stages, the day and night labels of the same kind of object are merged.
That is, the final outcome is still merged into three categories, vehicle, pedestrian, and rider.
With the proposed label-extension strategy, the deep-network model can focus on learning effective feature representations that can discriminate the dissimilar daytime and nighttime objects as two separated classes. 
The accuracy can then be considerably improved in our experience.
\label{method:labels_extension}

%Accordingly, if we train the model labeling the two objects with great differences as the same class, we will not acquire the model that performs robust detection of the objects between different scenes. According to the time period of the image, we can divide labels into daytime objects and nighttime objects, so that the model can focus on learning detection of objects in the identical time of day. On the training dataset, we expand the original three kinds of labels into six kinds of labels(vehicle, pedestrian, rider, night$\_$vehicle, night$\_$pedestrian and night$\_$rider). Finally, when calculating accuracy, we merge the labels of the night objects back into the corresponding labels of three kinds of objects.

% Figure

\subsubsection{Training Data Increments}
It is a common technique to enhance the accuracy by using more training samples. 
Accordingly, we extend the training data from an outside dataset, BDD100K\cite{yu2018bdd100k}. 
This dataset contains 10 categories including bus, light, sign, person, bike, truck, motor, car, train, rider. 
We remove the four categories, light, sign, bike and train, and reorganize the remain six into three categories so that they are coincide with the MMSP Competition.
To achieve this, we unify bus, truck, and car into the vehicle category, motor and rider into the rider category, and keep person as the pedestrian category, respectively.
The training and evaluation sets of BDD100K are combined with MMSP training set for model learning.

\subsection{Efficiency Improvement}
After applying the strategies depicted above, the mAP of Cascade  R-CNN  with  ResNeXt-101  backbone is boosted to 0.60 on the first stage public testing dataset, which is much higher than the original mAP of 0.40. However, the inference speed of the model is merely 0.17 FPS (frames per second) on NVIDIA TX2 (embedded computing device), which is non-satisfied yet. 
To further speed up the inference of the model, several techniques could be used.
For example, network quantization and filter pruning are common strategies that can be exploited. Although many methods can be utilized to compress a model, they are time-consuming to training a compact model while keeping the required accuracy. 
To boost the inference speed in a limited period, we follow two design principles that are easy to be realized. 
The first is to choose an efficient backbone network and the second is to re-size the input image to a smaller resolution.

\label{efficiency}
\subsubsection{Backbone Net}
In the above, we exploit ResNetx-101 FPN as the backbone network with the input resolution of 1920$\times$1080 pixels. 
As mentioned, the model ahcieves 0.6 mAP on the public testing dataset, but is computationally expensive and memory intensive. 
The first improvement is to reduce the complexity of the backbone network under a fixed input size, 1920$\times$1080 pixels.
We gradually replace the backbone network with the CNN models that have fewer parameters, but keeps the resulted detector meeting the required accuracy. 
We conduct our backbone-replacement evaluation on the ResNet series.
Our experiments reveal that employing RestNet-18 FPN as the backbone network can still achieve the mAP of 0.56 on the first stage public testing dataset. 
The inference speed is upgraded to 1.4 FPS.

\subsubsection{Input Resolution}
To additionally improve the inference speed of the above model, we adopt a straightforward strategy: reducing the input image resolution.
We re-size the input image to a smaller resolution before sending it to the model.
This step takes only a very limited time and thus does not affect the inference speed much. 
We can then manipulate only a single parameter, the re-scaling size, for speeding up the model. 
According to our experimental study, the input resolution of 960$\times$540 pixels can stil meet the minimum requirement of accuracy on the first stage public testing dataset, where the mAP on this resolution is 0.53, and the inference speed is boosted to 2.3 FPS.

Note that in the above, the inference speed (in terms of FPS) does not coincide with the officially announced results of MMSP Competiion (TABLE \ref{table:mmsp_result}).
It is because that the inference speed presented above does not include the loading time of testing images and deep-learning model from disk.

We have tried to use a lightweight object detector, Pelee~\cite{wang2018pelee}, in our study too. 
Pelee is more favorable than existing state-of-the-art computationally efficient models such as ShuffleNet \cite{zhang2018shufflenet} and MobileNet \cite{howard2017mobilenets}. 
However, when we evaluate the accuracy of Pelee model on the first stage public testing dataset, only 0.23 mAP is attained, and thus we have not chosen to explore it in this study.

% In our case, the Cascade R-CNN is chosen as the base detector. The backbone of the network is ResNet-18\cite{he2016deep}. According to the idea of Feature Pyramid Network, the intermediate features are extracted for better small object detection. Then the Cascade R-CNN, based on Faster R-CNN, will iteratively apply bounding box regression to refine the final results.
% The number of total parameters is 13,995,101. Stored in 32bit float, the size of the model is about 55 MB. The total operations (GOP/Frame) is 64,008,767,997, having an input image with 960 $\times$ 540. The processing time in Nvidia Jetson TX2 is around 0.025s per image.

\begin{table}[t]
\caption{Architecture of the ResNet-18 backbone networks.}
\begin{center}
\begin{tabular}{|c|c|c|}
\hline
Layer & \begin{tabular}[c]{@{}c@{}}Output\\ Size\end{tabular} & ResNet-18 \\ \hline
Input & 960$\times$540 & image \\ \hline
Conv1 & 480$\times$270 & 7$\times$7, 64, stride 2 \\ \hline
MaxPool & 240$\times$135 & 3$\times$3 max pool, stride 2 \\ \hline
Stage1 & 240$\times$135 & \begin{tabular}[c]{@{}c@{}}3$\times$3, 64\\ 3$\times$3, 64\\ $\times$2 \end{tabular} \\ \hline
Stage2 & 120$\times$68 & \begin{tabular}[c]{@{}c@{}}3$\times$3, 128\\ 3$\times$3, 128\\ $\times$2\end{tabular} \\ \hline
Stage3 & 64$\times$37 & \begin{tabular}[c]{@{}c@{}}3$\times$3, 256\\ 3$\times$3, 256\\ $\times$2\end{tabular} \\ \hline
Stage4 & 30$\times$17 & \begin{tabular}[c]{@{}c@{}}3$\times$3, 512\\ 3$\times$3, 512\\ $\times$2\end{tabular} \\ \hline
Complexity &  & 56 GOPs \\ \hline
Parameters &  & 11M \\ \hline
\end{tabular}
\end{center}
\label{fig:day_result}
\end{table}

\section{Experimental Results}
\label{experiments}

In this section, we evaluate the performance of our model on the MMSP Competition dataset. 
We also conduct ablation studies to evaluate our strategies.\\

\noindent\textbf{Dataset:} We evaluate the performance of our model on the dataset from
MMSP Competition. 
MMSP dataset contains 89,002 annotated 1920$\times$1080 images for training. 
During the competition, there are three kinds of testing sets including 1st-stage \textit{public testing dataset} (1,500 full HD test images), 2nd-stage \textit{testing dataset for qualification competition} (4,500 images), and 3rd-stage \textit{private testing dataset for final competition} (3,000 images). 
The objects in images are annotated with three categories: pedestrian, vehicle, and rider. During the competition, the ground truth labels are not provided. After the competition ends,  the annotations of these testing sets are available on the MMSP Competition website. In the following experiments, the accuracy of the model is evaluated on the private testing dataset for final competition.

% Figure
\begin{figure}[t]
  \begin{center}
    \includegraphics[width=0.8\columnwidth]{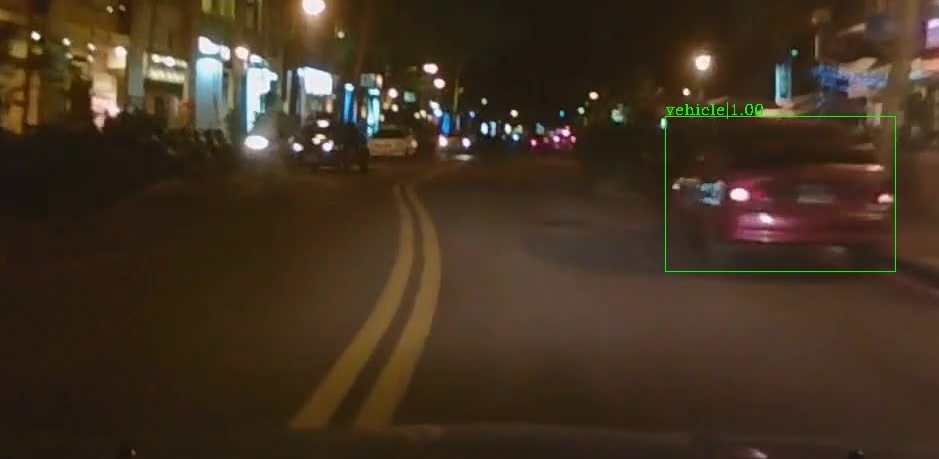}
  \end{center}
  \caption{The qualitative detection results of the model trained with three original labels (pedestrian, vehicle, and rider). The detector fails to recognize vehicles with obscure appearance in this example.}
  \label{fig:3labels_result}
  \begin{center}
    \includegraphics[width=0.8\columnwidth]{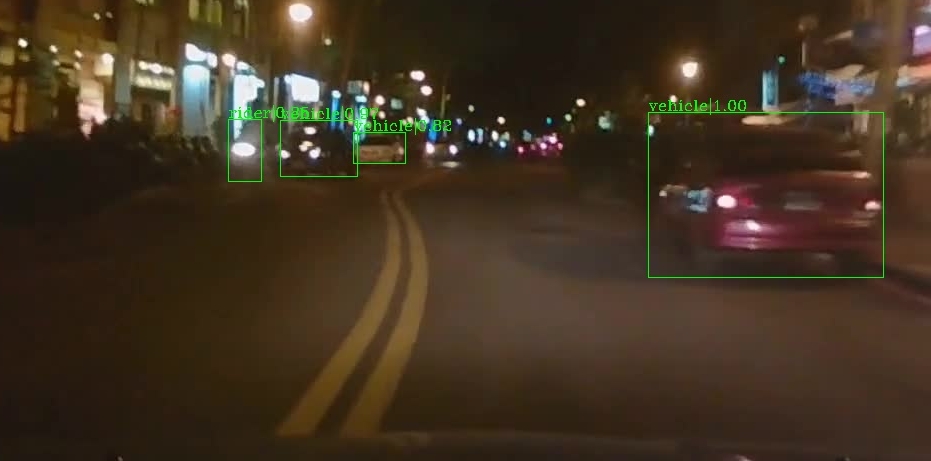}
  \end{center}
  \caption{The qualitative detection results of the model trained with the extended labels (adding another set of labels for objects in the nighttime). More vehicles in the nighttime can then be detected successfully.}
  \label{fig:6labels_result}
\end{figure}

% Figure
% \begin{figure}[ht]
%   \begin{center}
%     \includegraphics[width=0.8\columnwidth]{figures/6labels_example.jpg}
%   \end{center}
%   \caption{The qualitative detection results of the model trained with the extended labels (adding another set of labels for objects in the nighttime). More vehicles in the nighttime can then be detected successfully.}
%   \label{fig:6labels_result}
% \end{figure}

\subsection{Implementation Details}
We implement the model with an open-source object detection toolbox based on PyTorch, mmdetetion\cite{chen2019mmdetection}. Our model is trained end-to-end with 10 epochs. The learning rate is set to 0.02 at the beginning and then decayed by a factor of 0.1 at 80$\%$ of the total iterations. The four subnets of Cascade R-CNN are one RPN and three for detection with IoU @ $\left\{ 0.5, 0.6, 0.7 \right\}$, respectively.

\subsection{Results on the Competition Dataset}
The officially announced competition results include the evaluation of model size, computation complexity, and speed on NVIDIA Jetson TX2, respectively.  
Table \ref{table:mmsp_result} shows the final results of the teams that achieve the accuracy greater than 0.43 mAP on final testing dataset. 
Among them, the inference speed of our model gets 2nd place, while the number of model parameters of our model is the fewest. \\

\noindent\textbf{Improvement of Labels Extension}. As shown in Fig. \ref{fig:3labels_result}, when the original labels are adopted, the detector is struggled at detecting vehicles in the nighttime because of the strong appearance variations between nighttime and daytime. 
To address this problem, we extend labels as mentioned in \ref{method:labels_extension}. 
After the label extension, the detector can successfully recognize more vehicles in the nighttime. 
As shown in Fig. \ref{fig:6labels_result}, a rider and two vehicles on the left side of the image can be detected after the label-extension improvement.

\begin{table}[t]
\caption{Final Evaluation Result of MMSP 2019 Embedded Deep Learning Object Detection Model Competition. The speed is evaluated by the average execution time of the process to detect 3,000 testing images (including I/O time).}
\begin{center}
\begin{tabular}{@{}ccccc@{}}
\toprule
Team &  \begin{tabular}[c]{@{}c@{}}mAP\\ (IoU @ 0.5)\end{tabular} &  \multicolumn{1}{c}{\begin{tabular}[c]{@{}c@{}}Model Size\\ (MByte)\end{tabular}} &   \multicolumn{1}{c}{\begin{tabular}[c]{@{}c@{}}Complexity\\ (GOPS/frame)\end{tabular}} &
 \multicolumn{1}{c}{\begin{tabular}[c]{@{}c@{}}Speed\\ (ms/frame)\end{tabular}} \\ \midrule
R.JD & 0.538 & 124 & 43 & 460 \\
nctuai & 0.476 & 195 & 490 & 1338 \\
chenjiaqi & 0.461 & 114 & 339 & 1195 \\
\textbf{IMMVP (ours)} & 0.460 & \textbf{57} & 724 & 510 \\
NPUST-MIS & 0.439 & 238 & 115 & 514 \\ \bottomrule
\end{tabular}
\end{center}
\label{table:mmsp_result}
\end{table}

\subsection{Ablation Studies on Temporal and Spatial Resolution}
In the following, we explore the affection of sampling rate of frames to the accuracy on the final testing dataset. 
Then, we show the inference speed and the accuracy of the models with different image resolutions. 
%The following results are evaluated by our program. \\

\noindent\textbf{Affection of Frame Selection to the Accuracy}.
To figure out the affection of sampling rate to the accuracy on the final testing image set, we perform an ablation study with different frame selection rates. 
TABLE \ref{table:sample_rate} lists the accuracy of Cascade R-CNN with ResNet-18 FPN backbone employing different sample rate. 
The best accuracy in terms of mAP is achieved when we sample one frame from ten in a sequence.

\noindent\textbf{Affection different input resolution.} We evaluate the inference speed of our model with different input resolution on PC (NVIDIA Tesla V100 GPU), and TX2 (embedding device with NVIDIA Pascal CUDA GPU). The results are shown in TABLE IV. When the input resolution is 960$\times$540 and 1920$\times$1080, our model achieves both higher than 0.5 mAP.
Real-time detection is achievable on PC (with 24.3 FPS and 38.2 FPS, respectively).

Our mAPs shown above are all evaluated via VOC2012 mAP @ IoU 0.5 with the codes on the github https://github.com/yxlijun/Pelee.Pytorch/tree/master/data.
One may note that the testing accuracy achieved by our IMMVP model (Cascade R-CNN w/ResNet-18 FPN shown in Figure~\ref{fig:model_arc} with the input resolution 960$\times$540) is 0.507 on TABLE IV, which has a gap over 0.46, the officially announced mAP of our model on the final testing dataset (3,000 images) shown in TABLE \ref{table:mmsp_result}.
It could be because that the officially announced mAP is evaluated via MS COCO metric@IoU 0.5 but not VOC2012 mAP @ IoU 0.5 adopted for our ablation study.
According to our empirical testing with other datasets, the metrics of VOC2012- and MS COCO-mAP @ IOU 0.5 actually produce different evaluation results.
To align the accuracy to the official announcement, we evaluate our model via the on-line testing service of the official website and the results are given as follows. 
The accuracy of different sampling rates from 1 to 1/30 become 0.466, 0.502, 0.460, and 0.453, respectively, on TABLE \ref{table:sample_rate}, and that of the different input sizes from 1920$\times$1080 to 320$\times$180 become 0.511, 0.460, 0.377, and 0.341, respectively, on TABLE IV.
Nevertheless, all of our ablation studies are evaluated via the VOC2012 mAP @ IoU 0.5 codes mentioned above for a fair comparison.

\begin{table}[t]
\caption{Results on the final testing dataset with different sampling rates of training. Outside dataset is used in this experiment.}
\begin{center}
\begin{tabular}{@{}ccll@{}}
\toprule
Model & Resolution & \multicolumn{1}{c}{Sampling Rate} & $mAP_{voc}$        \\ \midrule
 & \multirow{3}{*}{960$\times$540} & 1          & 0.507        \\
\multirow{2}{*}{\begin{tabular}[c]{@{}c@{}}Cascade R-CNN w/   \\        ResNet-18 FPN\end{tabular}} &         & 1/10        & \textbf{0.533}       \\
 &                                & 1/20        & 0.512 \\
 &                                & 1/30        & 0.511        \\ \bottomrule
\end{tabular}
\end{center}
\label{table:sample_rate}
\end{table}

\begin{table}[t]
\label{table:input_size}
\begin{center}
\caption{Results of different resolutions on the final testing dataset. The inference speed(FPS) are evaluated on both PC (NVIDIA TESLA V100 GPU) and TX2 (NVIDIA Pascal GPU).}
\begin{tabular}{@{}cclll@{}}
\toprule
Model & Input & \multicolumn{1}{c}{$mAP_{voc}$} & PC & \multicolumn{1}{c}{TX2} \\ \midrule
 & 1920$\times$1080 & \textbf{0.546} & 24.3 & 1.4 \\
\multirow{2}{*}{\begin{tabular}[c]{@{}c@{}}Cascade R-CNN w/ \\ ResNet-18 FPN\end{tabular}} & 960$\times$540 & 0.507 & 38.4 & 2.3 \\
 & 640$\times$360 & 0.422 & 41.6 & 2.7 \\
 & 320$\times$180 & 0.403 & \textbf{45.4} & \textbf{3.2} \\ \bottomrule
\end{tabular}
\end{center}
\end{table}

\section{Conclusion}
\label{conclusion}
In this paper, we introduce useful strategies to improve the accuracy of the daytime and nighttime on-road object detector. 
First, we remove the redundant information of the training data via sample selection. 
We then propose to use label-extension to address the problem of large appearance variations between daytime and nighttime objects. 
Finally, the training set is expanded with related outside data. 
To accelerate the inference speed of the model while maintaining accuracy, we choose the ResNet-18 FPN as the backbone of the Cascade R-CNN. 
As a result, the size of the model is 57 MByte, with the inference speed of 2.3 FPS on NVIDIA Jetson TX2. 
It achieved 0.460 mAP (0.507 mAP by our evaluation) on the MMSP Competition final testing dataset.

\section*{Acknowledgment}
This work is supported in part by the Ministry of Science and Technology of Taiwan under grants MOST 108-2634-F-001-004.

{\small
\bibliographystyle{IEEEtran.bst}
\bibliography{IEEEfull.bib}
}

\end{document}